\documentclass[runningheads]{llncs}
\usepackage{graphicx}
\usepackage{booktabs}
\usepackage{array}
\usepackage[table,xcdraw]{xcolor}
\usepackage{multirow}
\usepackage{multicol}
\usepackage{cite}
\usepackage{amsmath}
\usepackage{dirtytalk}
\usepackage{xcolor}
\usepackage{hyperref}
\definecolor{darkgreen}{rgb}{0.0,0.5,0.0}
\definecolor{darkpurple}{rgb}{0.5,0.0,0.5}
\usepackage{amssymb}
\usepackage{pifont}
\newcommand{\cmark}{\ding{51}}%
\newcommand{\xmark}{\ding{55}}%
\usepackage{multirow}
\usepackage{multicol}
\usepackage{cite}
\usepackage{amsmath}
\usepackage{dirtytalk}
\usepackage{xcolor}
\usepackage{hyperref}
\definecolor{darkgreen}{rgb}{0.0,0.5,0.0}
\definecolor{darkpurple}{rgb}{0.5,0.0,0.5}
\usepackage{appendix}

\begin{document}
\title{MedPromptX: Grounded Multimodal Prompting for Chest X-ray Diagnosis}

\titlerunning{MedPromptX}

\author{Mai A. Shaaban \inst{1} \and
Adnan Khan \inst{2} \and
Mohammad Yaqub \inst{1}}

%
\authorrunning{Mai A. Shaaban et al.}
%
\institute{Mohamed bin Zayed University of Artificial Intelligence, Abu Dhabi, UAE
\\ \email{\{mai.kassem, mohammad.yaqub\}@mbzuai.ac.ae}\and
School of Computer Science, Carleton University, Ottawa, CA
\\ \email{adnankhan5@cmail.carleton.ca}}

\maketitle              
\begin{abstract}
Chest X-ray images are commonly used for predicting acute and chronic cardiopulmonary conditions, but efforts to integrate them with structured clinical data face challenges due to incomplete electronic health records (EHR). This paper introduces MedPromptX, the first clinical decision support system that integrates multimodal large language models (MLLMs), few-shot prompting (FP) and visual grounding (VG) to combine imagery with EHR data for chest X-ray diagnosis. A pre-trained MLLM is utilized to complement the missing EHR information, providing a comprehensive understanding of patients' medical history. Additionally, FP reduces the necessity for extensive training of MLLMs while effectively tackling the issue of hallucination. Nevertheless, the process of determining the optimal number of few-shot examples and selecting high-quality candidates can be burdensome, yet it profoundly influences model performance. Hence, we propose a new technique that dynamically refines few-shot data for real-time adjustment to new patient scenarios. Moreover, VG narrows the search area in X-ray images, thereby enhancing the identification of abnormalities. We also release MedPromptX-VQA, a new in-context visual question answering dataset encompassing interleaved images and EHR data derived from MIMIC-IV and MIMIC-CXR-JPG databases. Results demonstrate the SOTA performance of MedPromptX, achieving an 11\% improvement in F1-score compared to the baselines. Code and data are publicly available on \url{https://github.com/BioMedIA-MBZUAI/MedPromptX}.

\end{abstract}

\section{Introduction}
Emerging machine learning and deep learning advancements are assisting radiologists in detecting chest X-ray abnormalities, streamlining diagnostic processes \cite{najjar2023redefining,van2021does}. While traditional diagnosis based solely on imaging data can be effective, incorporating patients' clinical history can significantly improve diagnostic outcomes, underscoring the importance of multimodal approaches \cite{med-palm-m}. The integration of electronic health records (EHR) has been challenged by its inherent incompleteness \cite{shah2020secondary}. For instance, the missing values of normal ranges for laboratory tests can complicate the interpretation of medical datasets like MIMIC-IV \cite{mimic-iv}. Hence, processing textual descriptions of static EHR events can help complement the missing information by leveraging the capabilities of large language models pre-trained on large-scale medical data. We provide a subset from the MIMIC-IV dataset in Table~\ref{tab:ehr} that exemplifies undefined upper and lower thresholds for some lab tests. To this end, large language models (LLMs), as in \cite{zhu2024prompting}, have shown promise in clinical prediction by fine-tuning with prompts leveraging structured EHR data. While multimodal LLMs like BiomedGPT \cite{zhang2024biomedgpt} represent a major advancement in biomedical AI by handling various tasks across modalities and surpassing SOTA results, there is limited research on exploring the capabilities of LLMs with structured EHR data, particularly lab test results. Additionally, multimodal techniques, specifically visual grounding techniques, as explored in \cite{Ichinose_2023}, further exemplify progress in automating associations between image features and descriptive reports in CT imaging. Despite these advancements, there remains a gap in the integration of multimodal data for enhancing diagnostic accuracy in chest X-ray analysis \cite{li2023comprehensive}.

\begin{table}[htbp]
\centering
\caption{Example of incomplete information about lab test results in electronic health records.}
\label{tab:ehr}
\resizebox{0.6\textwidth}{!}{%
\begin{tabular}{@{}ccccc@{}}
\toprule
\rowcolor[HTML]{D9D9D9} 
\textbf{Value} & \textbf{Unit}  & \textbf{Label}                   & \textbf{Low} & \textbf{High} \\ \midrule
479   & mL    & Tidal Volume (observed) & 299 & 750  \\
9     & L/min & Minute Volume           & -   & 12   \\
39.9  & L/min & Flow Rate (L/min)       & -   & -    \\
24    & cmH2O & Plateau Pressure        & -   & 31   \\ \bottomrule
\end{tabular}%
}
\end{table}

Training LLMs or even fine-tuning can be computationally expensive \cite{med-llm-survey}. Therefore, a crucial breakthrough lies in few-shot prompting \cite{few-shot-learners}, which enables rapid adaptation to new diagnostic tasks with minimal labelled data and without parameter updates. This empowers medical professionals to efficiently use accurate diagnostic solutions tailored to specific patient cases \cite{med-flamingo}. In addition, few-shot prompting addresses the challenge of hallucination in LLMs, guiding the output and ensuring the reliability of diagnostic results \cite{hallucination}. Nevertheless, the quality and the quantity of the few-shot data play a pivotal role in influencing performance \cite{flamingo, few-shot-knowledge}. 


To this end, we introduce \textbf{MedPromptX} and a new multimodal in-context learning dataset. To the best of our knowledge, MedPromptX is the first model to integrate multimodal LLMs, few-shot prompting, and visual grounding for chest X-ray diagnosis. MedPromptX addresses the challenge of incomplete EHR by complementing missing information through a pre-trained multimodal LLM and focusing on relevant image regions through visual grounding. Additionally, we propose a dynamic proximity selection (DPS) technique that refines few-shot data in real-time. DPS involves analyzing a few examples of positively and negatively diagnosed patients. This technique allows the model to capture the nuanced relationships between patient history and patient outcomes, enhancing diagnostic accuracy while reducing the dependency on extensive labelled datasets, positioning our framework as a significant advancement in the field. Our main contributions are as follows:
\begin{itemize}
    \item Introducing \textbf{MedPromptX}, a novel diagnostic model for chest X-ray images that harnesses multimodal LLMs (MLLMs), few-shot prompting (FP) and visual grounding (VG), enabling more accurate prediction of abnormalities.
    \item Mitigating the incompleteness in EHR data by transforming inputs into a textual form, adopting pre-trained MLLMs. 
    \item Extracting the logical patterns discerned from the few-shot data efficiently by implementing DPS, allowing for the capture of the underlying semantics.
    \item Constructing MedPromptX-VQA, a new in-context learning dataset tailored for VQA with interleaved chest X-ray images and structured medical data.
\end{itemize}

\section{Methodology}

\begin{figure}[!t]
    \centering
    \includegraphics[width=\textwidth]{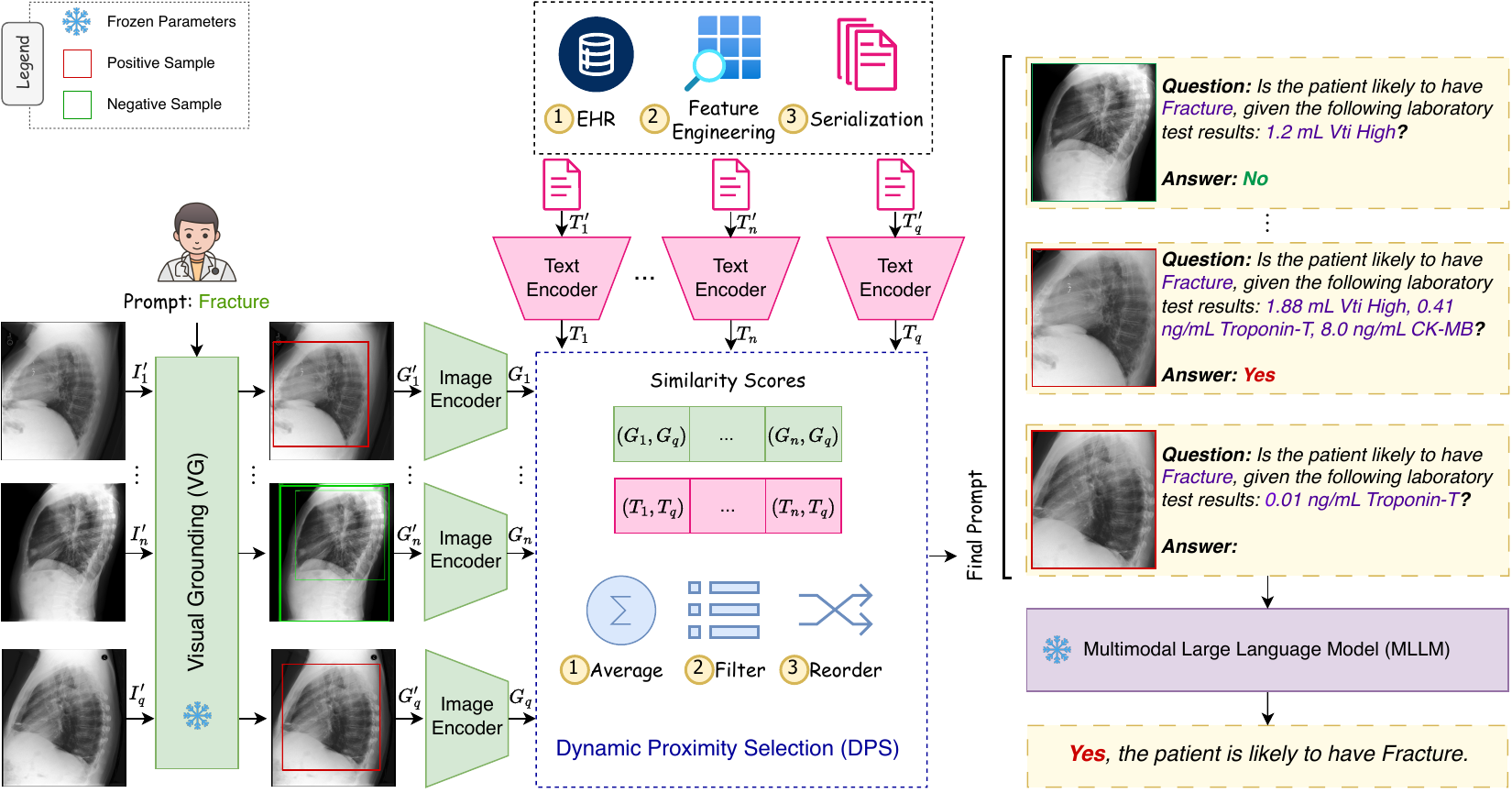}
    \caption{MedPromptX: each input sample consists of an image $I'$ and corresponding text $T'$ containing tabular features. (1) The \textcolor{darkgreen}{VG} model takes $I'$ and generates a grounded image $G'$ by prompting the desired output. (2) The grounded image embeddings $G$ and text embeddings $T$ of the candidates are processed by the \textcolor{blue!80!black}{DPS} technique to calculate their relevancy scores to a query sample $q$. (3) \textcolor{darkpurple}{MLLM} ingests a few-shot prompt and predicts whether a patient is likely to have a targeted disease.}
    \label{fig:method}
\end{figure}

\label{subsection:approach}
\subsection{MedPromptX for Diagnosis}
The workflow of MedPromptX in Figure~\ref{fig:method} can be conceptualized as a four-phase process. Let $\mathcal{C} = \{ (I'_1, T'_1), \ldots, (I'_n, T'_n) \}$ denotes a set of $n$ candidates and $q = (I'_q, T'_q)$ denotes the query sample. $I'$ is a chest X-ray image and $T'$ is the corresponding text containing EHR data. First, the \textcolor{darkgreen}{visual grounding (VG)} module crops irrelevant parts in each sample and generates grounded image $G'$ by prompting the class. Then, frozen image and text encoders generate image $G$ and text $T$ embeddings, respectively. Next, the \textcolor{blue!80!black}{dynamic proximity selection (DPS)} module refines the candidates, resulting in $\mathcal{E}$, where $\mathcal{E} \subseteq \mathcal{C}$. Finally, the \textcolor{darkpurple}{multimodal large language model (MLLM)} ingests the final prompt containing a reordered subset $\mathcal{E}$ to predict the abnormality in query patient $q$.

\subsubsection{Visual Grounding (VG)}
Conventional object detection methods often confront limitations regarding their capacity to recognize predefined classes of objects \cite{glip,gdino}. Integrating new classes into these models necessitates an exhaustive process of data collection, annotation and model retraining. We use Grounding DINO (GDINO) \cite{gdino} (the \textcolor{darkgreen}{VG} component in Figure~\ref{fig:method}) to address this challenge by detecting arbitrary objects in real-time delineated through human language inputs, a concept commonly referred to as zero-shot detection.

GDINO uses DINO \cite{zhang2022dino}, a SOTA transformer-based object detection algorithm, with GLIP \cite{glip} pre-training that focuses on grounding textual descriptions to visual elements in a given image. GDINO is a two-stream framework where multi-scale image and text features are extracted separately using backbone architectures such as Swin Transformer \cite{swin} and BERT \cite{bert}, respectively. These features are then transformed into a unified representation space through multiple layers of feature enhancers, incorporating deformable self-attention for image features and regular self-attention for text features.

To detect visual evidence (i.e., grounded image) denoted as $G'$, a clinician passes a textual input $e$ of a pathological condition (e.g., \textit{Pneumonia}) along with an X-ray image $I'$ to the VG model. The model assigns scores to particular regions based on their prominence in the image $VG(I',e)=\{p(G'_1),\dots,p(G'_k)\}$, where $k$ is the total number of detected regions and $p$ is the score. We then consider $G'$ with the highest score for the subsequent phases of our model. This approach narrows the search area in an X-ray image. 

\subsubsection{Dynamic Proximity Selection (DPS)}
The performance of FP is highly sensitive to the design of the prompt. This includes the choice of examples, their order, and how well they align with the desired task. Misleading, ambiguous, or poorly chosen examples can lead to suboptimal or entirely incorrect outputs \cite{few-shot-knowledge,flamingo}. The \textcolor{blue!80!black}{DPS} method leverages a distance function $d$, such as cosine similarity to order candidate instances $\mathcal{C} = \{ (G'_1, T'_1), \ldots, (G'_n, T'_n) \}$, based on their proximity to a query instance $q$. Applying a similarity threshold dynamically filters out noisy candidates, enhancing the robustness and adaptability of the FP technique. Thus, the number of $n$ candidate samples can be reduced ($n-1$,$n-2$,\dots,1). Mathematically, the approach can be represented as:

\begin{equation}
\text{DPS}(\mathcal{C}, q) = \left\{ \frac{d({G_{c},G_q}) + d({T_{c},T_q})}{2} \geq th \right\}_{c \in \mathcal{C}}
\end{equation}

The result is a refined subset $\mathcal{E}$ where each candidate has a similarity score greater than or equal to a threshold $th$. In this method, an instance ${c \in \mathcal{C}}$ can be decomposed into either grounded image embeddings $G_c$ or text embeddings $T_c$ containing laboratory test results of a patient. After computing the similarity scores for text and images separately, the final score is obtained by averaging the scores from both modalities. Motivated by \cite{few-shot-knowledge,flamingo}, DPS positions the most closely related candidate directly before the query instance rather than allocating it at a greater distance. This order enhances the precision of the FP process.

\subsubsection{Multimodal LLM (MLLM)}
Incorporating descriptive information about clinical events can provide valuable context for understanding the reasoning behind model predictions, unlike classical machine learning algorithms, which treat input as numerical attributes without considering the semantic meaning. There are limited examples of open-source models that can ingest FP with interleaved modalities. One notable model is Med-Flamingo \cite{med-flamingo}, which has undergone pre-training on a vast array of medical data. Therefore, Med-Flamingo, which is based on Flamingo \cite{flamingo}, serves as the \textcolor{darkpurple}{MLLM} component in Figure~\ref{fig:method}. The Flamingo \cite{flamingo} framework can process inputs consisting of both textual and visual content and produce coherent textual output. Flamingo adopts a strategy of freezing the language model and vision encoder weights and establishing connections through learnable architectures. The key component is the perceiver resampler module, introduced in Flamingo to convert spatiotemporal features from the vision encoder into a fixed-size set of visual tokens, facilitating their integration into the language model's processing pipeline. Additionally, cross-attention layers are inserted between pre-trained language model layers, enabling the model to incorporate visual cues for tasks such as next-token prediction. The pivotal aspect of Flamingo is that it predicts the likelihood of text sequences $y$ when conditioned on accompanying images $x$ as follows:

\begin{equation}
    p(y | x) = \prod_{\ell=1}^{L} p(y_{\ell} | y_{<\ell}, x_{\leq\ell}).
\end{equation}

\noindent The notation \( y_{\ell} \) represents the \( \ell \)-th token in the sequence of $L$ language tokens constituting our input text, while \( y_{<\ell} \) denotes all preceding language tokens, and \( x_{\leq\ell} \) symbolizes the corresponding sequence of images.

\subsection{MedPromptX-VQA Dataset}
\label{subsection:dataset}
\begin{figure}[t!]
    \centering
    \includegraphics[width=\textwidth]{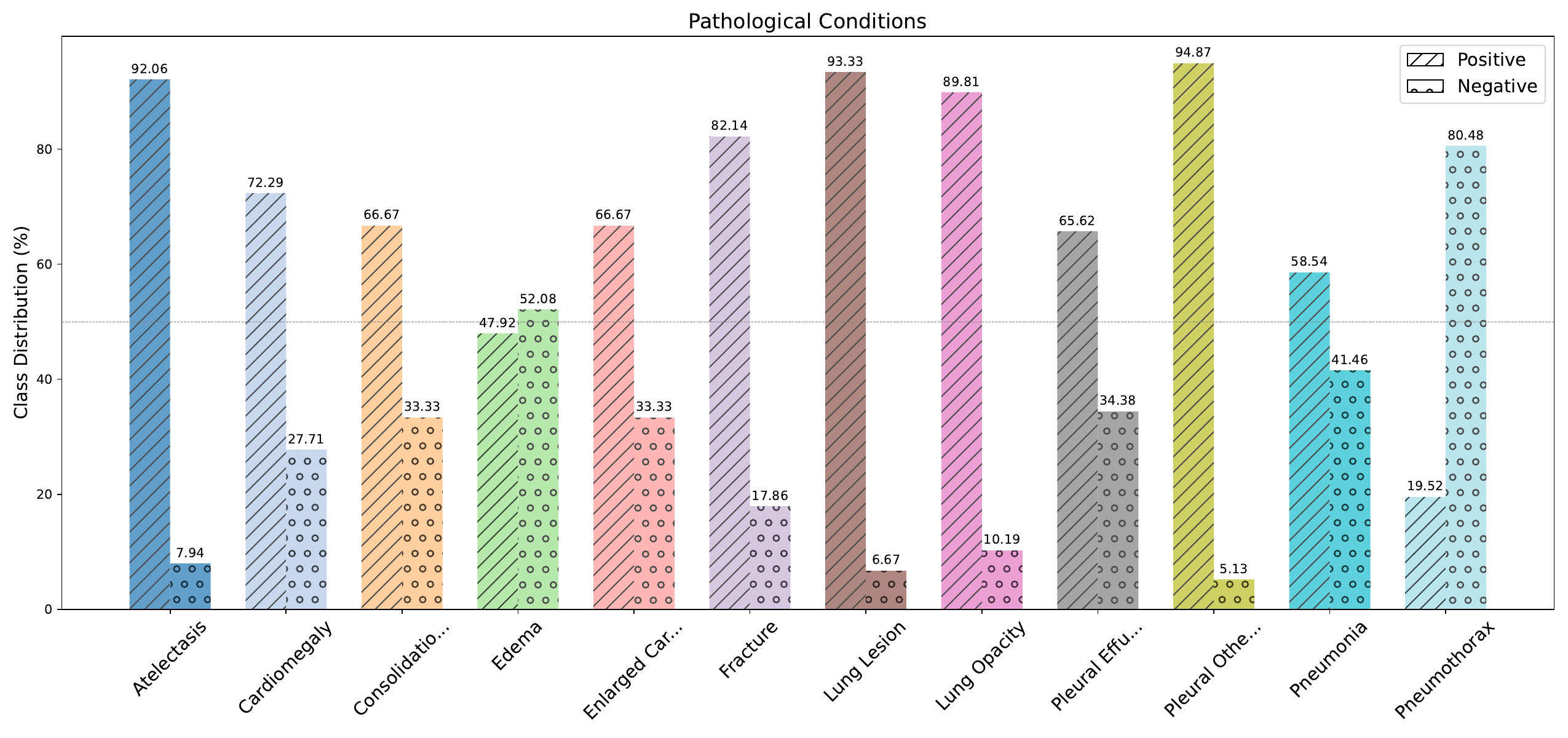}
    \caption{The ``Positive'' and ``Negative'' representations for 12 pathological conditions.}
    \label{fig:class}
\end{figure}
Our methodology involves constructing the MedPromptX-VQA dataset derived from a unified multimodal dataset, denoted as HAIM-MIMIC-MM \cite{haim}. HAIM-MIMIC-MM is a fusion of information sourced from MIMIC-IV \cite{mimic-iv} and MIMIC-CXR-JPG \cite{mimic-cxr} databases, meticulously curated to focus solely on patients with at least one chest X-ray procedure. This dataset encapsulates records from 7,279 hospitalization stays involving 6,485 distinct patients, thereby establishing a multimodal link encompassing tabular, textual and visual representations of patient health data.

In MedPromptX-VQA, patients are labelled with 12 pathological conditions: \textit{Atelectasis, Cardiomegaly, Consolidation, Edema, Enlarged Cardiomediastinum, Fracture, Lung Lesion, Lung Opacity, Pleural Effusion, Pleural Other, Pneumonia and Pneumothorax}. To alleviate the challenges of limited context length and hallucination in LLMs \cite{hallucination}, we transformed these labels into a binary single-label classification framework to ensure that the input fits the context length and to acquire a controlled output, rather than acquiring an open set of possible diagnoses. For each label, if a patient exhibits the condition, the corresponding label is assigned the value 1; otherwise, it is given the value 0. For this study, we specifically selected patients diagnosed with the aforementioned conditions, resulting in 968 records split into 501 positive and 467 negative samples. Figure~\ref{fig:class} shows the representations of the labels in the final dataset.

The creation of MedPromptX-VQA involves three steps: (1) extraction of laboratory test results from the \textit{chartevents} table within the MIMIC-IV dataset \cite{mimic-iv}, resulting in 357 features in total, (2) feature engineering, which includes identification of the most strongly correlated features in relation to the label using Pearson method, and (3) transformation of clinical charts into textual representations using comma-separated values (i.e., serialization). Finally, the dataset is structured to support the in-context learning task, where each record has interleaved images and texts, encompassing both positive and negative samples of patients. The motivation for feature selection is to maintain input consistency between the few-shot data and the query sample. This means that the features present in the query sample should already be represented by the candidates while also adhering to the context length. Hence, we set a maximum of 10 features per label. The selected features are presented in Table~\ref{tab:featues} in \textit{Appendix}.

\section{Experimental Setup} We employed a randomized order strategy for the input sequences across several SOTA models, namely Med-Flamingo \cite{med-flamingo}, OpenFlamingo \cite{open-flamingo}, BioMedLM \cite{biomedlm} and Clinical-T5-Large \cite{clinicalt5}. Moreover, the number of few-shot samples remained consistent at 6 across all the models and they were chosen randomly. For MedPromptX, the number of few-shot candidates is dynamically reduced by the DPS technique, resulting in a unique configuration for each query instance. The exclusion criteria for a candidate involve eliminating instances where the cosine similarity falls below a certain threshold, set at 70\%. Furthermore, all experiments were conducted using NVIDIA A100-SXM GPU equipped with 40GB of dedicated memory. For MedPromptX, the frozen language encoder employed is LLaMA-7B \cite{llama}, while the frozen visual encoder is CLIP ViT-L-14 \cite{vit}. Table~\ref{tab:models} in \textit{Appendix} shows detailed descriptions of the used models.

The prompt design for each model differs based on its capability. Accordingly, Med-Flamingo and OpenFlamingo ingest interleaved images and texts, excluding EHR data, whereas BioMedLM and Clinical-T5 use text, including EHR data. MedPromptX stands out as the sole model that processes interleaved grounded or original images and EHR text prompts. Below are examples for each type:

\begin{itemize}
    \item Image, Text: \say{\textless \textit{image}\textgreater \textit{Question:} Is the patient likely to have Cardiomegaly?}
    \item EHR Text: \say{\textit{Question:} Is the patient likely to have Cardiomegaly, given the following laboratory test results: 0.52 sec QTc?}
    \item Image, EHR Text: \say{\textless \textit{image}\textgreater \textit{Question:} Is the patient likely to have Cardiomegaly, given the following laboratory test results: 0.52 sec QTc?}
\end{itemize}

\section{Results and Discussion}

The results in Table~\ref{tab:few_shot} emphasize the complex nature of medical diagnosis, wherein multiple data modalities can provide complementary information leading to better model performance. The combination of imaging data with clinical text via MedPromptX seems significant in providing the model with a richer context, leading to more informed predictions. However, initial attempts yielded lower results, emphasizing the challenges in effectively integrating diverse data sources. With the implementation of DPS and VG, subsequent improvements were observed, suggesting that these strategies are crucial in overcoming the obstacles encountered when processing complex prompts. Although some clinical features may not significantly impact the final decision, refining the prompt using DPS could limit the ambiguity. However, it is also recommended to explore strategies such as causal inference in the future.

\begin{table}[t]
\centering
\caption{Performance of MedPromptX against SOTA baselines. Without DPS, candidate prompts in the 6-shot setting are randomly ordered. In contrast, with DPS, the ordering is determined by cosine similarity scores between the embeddings of each candidate and the test prompt, potentially reducing the number of candidates per record given a threshold. When VG is activated, the model processes images with contextual grounding. Conversely, when VG is deactivated, the model ingests original images.}
\label{tab:few_shot}
\resizebox{\textwidth}{!}{%
\begin{tabular}{@{}c|c|c|cccc@{}}
\toprule
\rowcolor[HTML]{D9D9D9} 
\textbf{Model}                                                       & \textbf{DPS Setting}                               & \textbf{VG Setting}        & \textbf{Precision} & \textbf{Recall} & \textbf{F1-score} & \textbf{Accuracy} \\ \midrule
BioMedLM                                                             &                                                    &                            & 0.665              & 0.210           & 0.484             & 0.536             \\
Clinical-T5-Large                                                    & \multirow{-2}{*}{\xmark}                         & \multirow{-2}{*}{N/A}      & 0.707              & 0.371           & 0.576             & 0.595             \\
Med-Flamingo                                                         &                                                    &                            & 0.545              & 0.220           & 0.461             & 0.501             \\
OpenFlamingo                                                         & \multirow{-2}{*}{\xmark}                         & \multirow{-2}{*}{\xmark} & 0.523              & 0.291           & 0.476             & 0.496             \\ \midrule
\rowcolor[HTML]{E7E6E6} 
\cellcolor[HTML]{E7E6E6}                                             & \cellcolor[HTML]{E7E6E6}                           & \xmark                   & 0.520              & 0.381           & 0.493             & 0.498             \\
\rowcolor[HTML]{E7E6E6} 
\cellcolor[HTML]{E7E6E6}                                             & \multirow{-2}{*}{\cellcolor[HTML]{E7E6E6}\xmark} & \cmark                    & 0.511              & 0.379           & 0.486             & 0.491             \\
\rowcolor[HTML]{E7E6E6} 
\cellcolor[HTML]{E7E6E6}                                             & \cellcolor[HTML]{E7E6E6}                           & \xmark                   & \underline{0.708}        & \textbf{0.581}  & \underline{0.658}       & \underline{0.659}       \\
\rowcolor[HTML]{E7E6E6} 
\multirow{-4}{*}{\cellcolor[HTML]{E7E6E6}\textbf{MedPromptX (ours)}} & \multirow{-2}{*}{\cellcolor[HTML]{E7E6E6}\cmark}  & \cmark                    & \textbf{0.773}     & \underline{0.565}     & \textbf{0.686}    & \textbf{0.689}    \\ \bottomrule 
\end{tabular}%
}
\end{table}

DPS enhances the model's ability to learn from limited data by reducing the number of ambiguous examples to 4 on average, contributing to better understanding. On the other hand, a random configuration of FP may introduce unintended biases or result in irrelevant guidance for the model. Moreover, the activation of VG empowers the model to restrict the search area within an image by generating output embeddings that encode semantic information instead of dealing with raw pixel data.

The performance gap observed when using VG solely may be attributed to training the VG model on general domain data rather than on chest X-ray images, particularly in handling the complexity inherent in cases where abnormalities are present in small regions. Providing additional context could bridge the gap, which was achieved by refining EHR data with DPS alongside VG.


\subsection{Ablations}


\begin{table}[t]
\centering
\caption{Comparing model performance using different number of instances for DPS initialization. The threshold is set at 0.7, and VG is enabled.}
\label{tab:shots}
\resizebox{0.6\textwidth}{!}{%
\begin{tabular}{@{}c|cccc@{}}
\toprule
\rowcolor[HTML]{D9D9D9} 
\textbf{Prompt   Setting} & \textbf{Precision} & \textbf{Recall} & \textbf{F1-score} & \textbf{Accuracy} \\ \midrule
4-shot                    & 0.640              & \underline{0.565}           & 0.609             & 0.609             \\
6-shot                    & \underline{0.773}              & \underline{0.565}          & 0.686             & 0.689             \\
8-shot                    & \textbf{0.789}              & 0.556           & 0.689             & 0.693             \\
10-shot                   & 0.732              & 0.541           & \underline{0.690}             & \underline{0.705}             \\
12-shot                   & 0.735              & \textbf{0.654}           & \textbf{0.733}             & \textbf{0.740}             \\ \bottomrule
\end{tabular}%
}
\end{table}



Initializing DPS with an increased number of shots provides the model with a broader range of context and examples to learn from, enabling it to generalize more effectively, as shown in Table~\ref{tab:shots}. However, the 6-shot setting strikes a balance between performance and ensuring the inclusion of all classes. In contrast, using a higher number of examples would necessitate dropping classes with insufficient positive or negative examples. Moreover, zero-shot assessment was unattainable due to the hallucination of the models, giving entirely incorrect output for some patient cases. This underscores the necessity for employing FP.

Adjusting the threshold for DPS can significantly affect performance; an extremely high threshold restricts the model from including meaningful examples, while an extremely low threshold retains nearly the same examples as in Figure~\ref{fig:threshold}. Moreover, the utilization of multimodal similarity, as shown in Table~\ref{tab:dps_modality}, enhances the instance selection process by capturing a more comprehensive representation of the data compared to single-modality approaches. 

\section{Conclusion}
This paper introduced MedPromptX, a novel prompting strategy that integrates clinical history with imaging data to support decision-making by clinicians for more accurate chest X-ray diagnosis. MedPromptX demonstrated an approach to address the challenges associated with medical data incompleteness, adaptability to new patient cases with limited labelled data and abnormality detection in X-ray images. Nevertheless, further improvements could be obtained by using fine-tuned backbones, which is beyond the scope of this study. Future work can include the accessibility of diverse and well-annotated datasets. Additionally, rigorous clinical trials and real-world deployment are necessary to validate the framework’s real-world effectiveness and clinical utility.

\bibliographystyle{splncs04}
\bibliography{ref}

\clearpage
\appendix
\counterwithin{figure}{section}
\counterwithin{table}{section}
\section{Appendix}

\begin{table}[htbp!]
\centering
\caption{Overview of large language models and visual-language models.}
\label{tab:models}
\resizebox{0.75\textwidth}{!}{%
\begin{tabular}{@{}c|c|c|c|c@{}}
\toprule
\rowcolor[HTML]{D9D9D9} 
\textbf{Model}    & \textbf{Pre-training Data}                     & \textbf{Visual Encoder} & \textbf{Language Model} & \textbf{Size} \\ \midrule
BioMedLM          & The Pile                                       &                         & Standard GPT-2          & 2.7B          \\
Clinical-T5-Large & MIMIC-III and MIMIC-IV                         & \multirow{-2}{*}{N/A}   & T5-Large                & 0.8B          \\
Med-Flamingo      & MTB and PMC-OA                                 & CLIP ViT-L-14           & LLaMA-7B                & 8.3B          \\
OpenFlamingo      & LAION-2B and Multimodal C4 & CLIP ViT-L-14           & MPT-1B                  & 3.0B            \\ \bottomrule
\end{tabular}%
}
\end{table}

\begin{figure}[]
    \centering
    \includegraphics[width=0.6\textwidth]{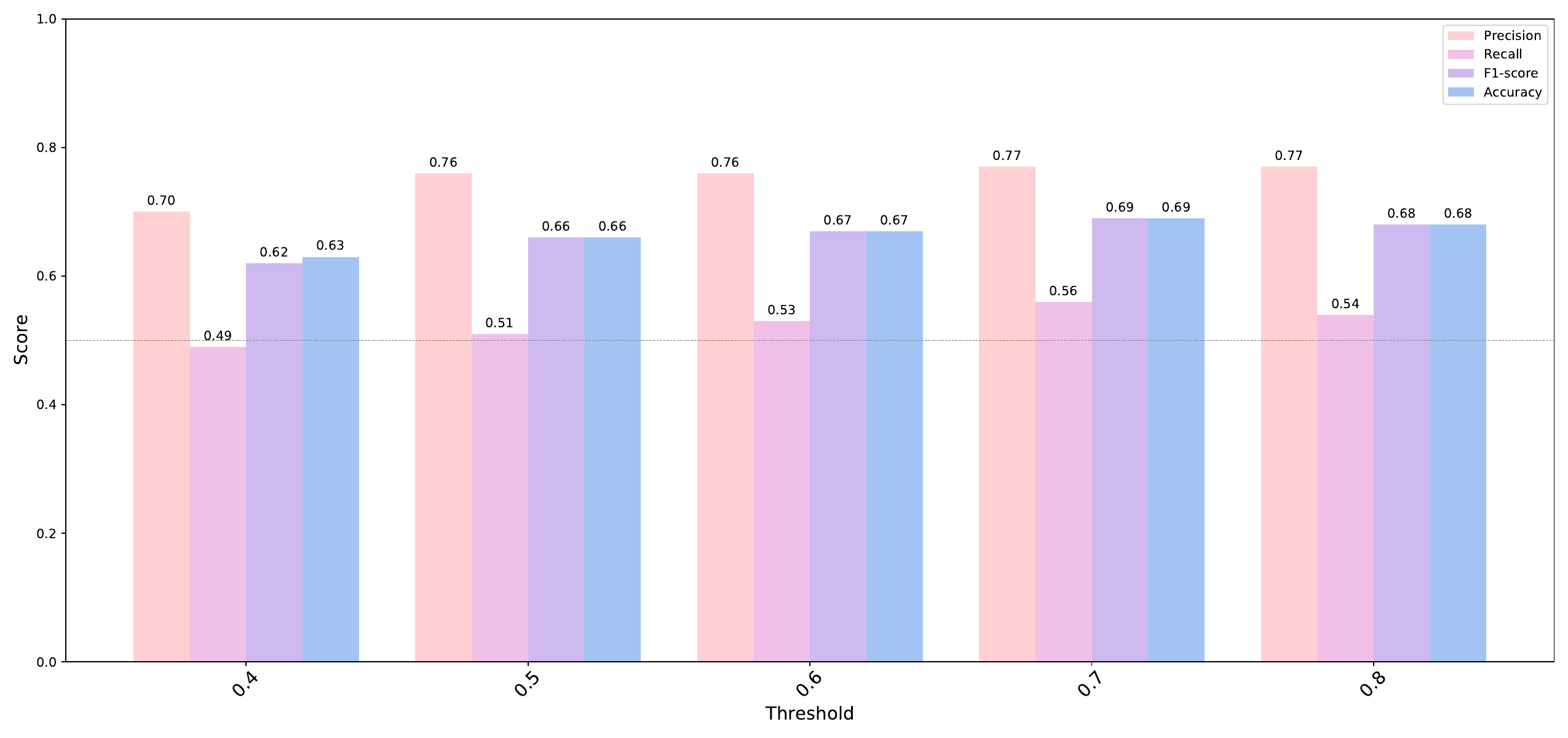}
    \caption{Comparison of MedPromptX under different DPS thresholds.}
    \label{fig:threshold}
\end{figure}

\begin{table}[t]
\centering
\caption{Comparison of employing DPS with averaged similarity scores from two modalities versus employing similarity scores based on a single modality.}
\label{tab:dps_modality}
\resizebox{0.6\textwidth}{!}{%
\begin{tabular}{@{}c|cccc@{}}
\toprule
\rowcolor[HTML]{D9D9D9} 
\textbf{DPS   Modality} & \textbf{Precision} & \textbf{Recall} & \textbf{F1-score} & \textbf{Accuracy} \\ \midrule
Text                    & 0.558              & 0.391           & 0.518             & 0.525             \\
Image                   & 0.748              & 0.463           & 0.632             & 0.642             \\
Multimodal              & \textbf{0.773 }             & \textbf{0.565}  & \textbf{0.686}    & \textbf{0.689}    \\ \bottomrule
\end{tabular}%
}
\end{table}

\begin{table}[htbp!]
\centering
\caption{Summary of the top correlated features that contribute to each label's prediction, providing a clear understanding of the significant variables.}
\label{tab:featues}
\resizebox{0.8\textwidth}{!}{%
\begin{tabular}{@{}c|c|>{\centering\arraybackslash}m{0.7\textwidth}@{}}
\toprule
\rowcolor[HTML]{D9D9D9} 
\textbf{Label}               & \textbf{No. Features} & \textbf{Top Features}                                                                                                                                                                                                                     \\ \midrule
Atelectasis                  & 10                    & CO (Arterial), HDL,   Cholesterol, ELWI (PiCCO), T Low (APRV), GEDI (PiCCO), LDL measured, T High   (APRV), LDL calculated, Serum Osmolality                                                                                                           \\ \hline
Cardiomegaly                 & 10                    & BiPap bpm (S/T -Back   up), LDL measured, ELWI (PiCCO), D-Dimer, Impaired Skin Length \#5, Impaired   Skin Width \#5, Uric Acid, GEDI (PiCCO), QTc, Cholesterol                                                                                        \\ \hline
Consolidation                & 9                     & Manual Blood Pressure   Diastolic Right, Manual Blood Pressure Systolic Right, ELWI (PiCCO), GEDI   (PiCCO), CFI (PiCCO), Manual Blood Pressure Diastolic Left, Negative Insp.   Force, Cholesterol, PCA basal rate (mL/hour)                          \\ \hline
Edema                        & 10                    & SV (Arterial), CO   (Arterial), ELWI (PiCCO), CFI (PiCCO), GEDI (PiCCO), Bladder Scan Estimate,   SVV (Arterial), Gentamicin (Random), LDL measured, BiPap bpm (S/T -Back up)                                                                          \\ \hline
Enlarged   Cardiomediastinum & 5                     & ELWI (PiCCO), GEDI   (PiCCO), SVV (Arterial), D-Dimer, RCexp (Measured Time Constant)                                                                                                                                                                  \\ \hline
Fracture                     & 10                    & Absolute Count -   Monos, CK-MB, Absolute Count - Neuts, Troponin-T, CO2 production,   Differential-Bands, Vti High, Absolute Count - Lymphs, Chloride (whole   blood), Total Bilirubin                                                                \\ \hline
Lung Lesion                  & 10                    & Temporary Ventricular   Sens Setting mV, Temporary Venticular Sens Threshold mV, PCV Level, Absolute   Count - Neuts, GI \#1 Tube Mark (CM), Temporary Pacemaker Rate, Glucose (whole   blood), Ionized Calcium, Total Bilirubin, Absolute Count - Eos \\ \hline
Lung Opacity                 & 6                     & Cardiac Output   (thermodilution), Bladder Scan Estimate, Ammonia, Serum Osmolality, PBP   (Prefilter) Replacement Rate, Current Goal                                                                                                                  \\ \hline
Pleural   Effusion           & 10                    & SV (Arterial), CO   (Arterial), ELWI (PiCCO), Permanent Pacemaker Rate, GEDI (PiCCO), Gentamicin   (Random), SVV (Arterial), Arctic Sun/Alsius Temp \#2 C, Feeding Weight, Arctic   Sun/Alsius Temp \#1 C                                              \\ \hline
Pleural Other                & 10                    & PCV Level, Impaired   Skin Length \#2, Temporary Ventricular Sens Setting mV, Temporary Venticular   Stim Threshold mA, Temperature Celsius, Temporary Venticular Sens Threshold   mV, Troponin-T, Total Bilirubin, Mixed Venous O2\% Sat, PeCO2       \\ \hline
Pneumonia                    & 9                     & ELWI (PiCCO),   Recruitment Duration, T Low (APRV), CO (PiCCO), HDL, Cholesterol, SV   (Arterial), Impaired Skin Width \#5, LDL measured                                                                                                               \\ \hline
Pneumothorax                 & 10                    & HDL, Impaired Skin   Width \#3, Cardiac Output (thermodilution), Temporary Venticular Stim   Threshold mA, TCO2 (calc) Venous, Total Bilirubin, Tidal Volume (set), Venous   CO2 Pressure, Differential-Monos, Absolute Count - Eos                    \\ \bottomrule
\end{tabular}%
}
\end{table}

\end{document}